# Towards Earlier Detection of Oral Diseases On Smartphones Using Oral and Dental RGB Images

Ayush Garg[1], Julia Lu[2], and Anika Maji[3]

[1]Dublin High School, ayushgargemail@gmail.com
[2]Homestead High School, julia.m.lu@gmail.com
[3]The Harker School, anika.maji@gmail.com

## ABSTRACT

Oral diseases such as periodontal (gum) diseases and dental caries (cavities) affect billions of people across the world today. However, previous state-of-the-art models have relied on X-ray images to detect oral diseases, making them inaccessible to remote monitoring, developing countries, and telemedicine. To combat this overuse of X-ray imagery, we propose a lightweight machine learning model capable of detecting calculus (also known as hardened plaque or tartar) in RGB images while running efficiently on low-end devices. The model, a modified MobileNetV3-Small neural network transfer learned from ImageNet, achieved an accuracy of 72.73% (which is comparable to state-of-the-art solutions) while still being able to run on mobile devices due to its reduced memory requirements and processing times. A ResNet34-based model was also constructed and achieved an accuracy of 81.82%. Both of these models were tested on a mobile app, demonstrating their potential to limit the number of serious oral disease cases as their predictions can help patients schedule appointments earlier without the need to go to the clinic. *Source code link*[1].

## I. INTRODUCTION

Oral diseases, including periodontal, or gum, disease and dental caries, or cavities, affect about 45% of the global population despite being preventable if detected and treated early on (World Health Organization). Dental caries, or tooth decay, is caused by plaque, a coating of bacteria that feeds on sugars and refined carbohydrates, and can lead to infection of the nerve at the center of the tooth as well as considerable tooth damage (John Hopkins Medicine). Similarly, periodontal disease is an infection that affects the gum tissue around the teeth and usually occurs due to an accumulation of calculus, also known as hardened plaque or tartar, which may result in tooth mobility or eventual tooth loss (National Institute of Dental and Craniofacial Research). With about 19% of adults worldwide experiencing severe gum disease (World Health Organization) and nearly 50% of people globally affected by dental caries (World Health Organization), oral disease is a widespread problem. As oral diseases are mostly treatable if detected in their initial stages, the need for accessible and reliable oral diagnostic care outside of the clinic is apparent.

In the past, studies have focused on diagnosing oral diseases through X-ray imagery (Mao et al.; Chen et al.; Prajapati et al.), and have achieved up to a 95% accuracy on detecting periodontitis (Mao et al.). However, methods utilizing imaging modalities such as X-rays are inaccessible outside of the clinic and require specialized equipment to be useful. Research has also been done in the detection of these diseases through RGB imagery. In a 2023 study by Park et al., the authors classified periodontal diseases from a dataset of 220 RGB images. Using their model, they were able to classify images with calculus or inflammation with an accuracy of 75.54%, increasing classification accuracy by 11.45% when compared to a previous ResNet152 model (Park et al.). However, the authors' use of a custom Convolutional Neural Network (CNN) prevents its use on low-powered devices such as mobile phones and thus limits its accessibility to consumers. Recently, another study detected and classified 7 different oral diseases, including periodontal disease and dental plaque (Liu et al.). Their model, a MASK R-CNN deep learning model, achieved an accuracy of 90%. The team also proposed an iHome smart dental Health-IoT system connecting patients and hospitals for quick diagnoses. They implemented their model in several clinics, reducing the diagnosis time by an average of 37.5% and increasing the number of patients treated by 18.4%. (Liu et al.). However, the proposed iHome smart dental Health-IoT system requires an internet connection to work, which some developing regions lack (Poushter). Additionally, the system requires consumers to send images to the hospital for processing, forcing them to share data with the hospital and posing a potential security risk.

---

[1]https://github.com/megargayu/dentalclassification/

Other fields have had development in the use of lightweight machine learning models run on mobile phones. Li et al. created a deep learning system based on mobile fundus imagery for glaucoma detection in a 2020 paper, achieving an AUC of 87.3%. In addition, Reshan et al. used the MobileNet model to detect pneumonia from chest x-rays with the accuracy of 94.23%. A 2022 paper by Maduranga and Nandasena developed a mobile app to diagnose skin disease from dermatoscopic images using MobileNet and achieved an accuracy of 85%.

Our contributions are as follows:

(1) A MobileNet (Howard et al.) model trained on RGB intraoral images that can be used in smartphones for easier access and the use of non-specialized equipment in the diagnostic process. Using images provided by Park et al.'s paper, we trained a MobileNetV3-Small based model that generated an accuracy of 72.73%, recall of 100.00%, F1 score of 80.00%, and precision of 66.67%, while still being feasible to run on mobile devices.

(2) A ResNet34 model (He et al.) reaching an accuracy of 81.82%, recall of 75.00%, F1 score of 81.82%, and precision of 90.00%. This model also ran on the mobile device, albeit not as fast or efficient as the MobileNet model.

(3) A mobile app capable of attaining the same accuracy as the original MobileNet and ResNet34 models, demonstrating the viability of the model running on mobile devices.

## II. METHODS

### A. Datasets and Data Filtering

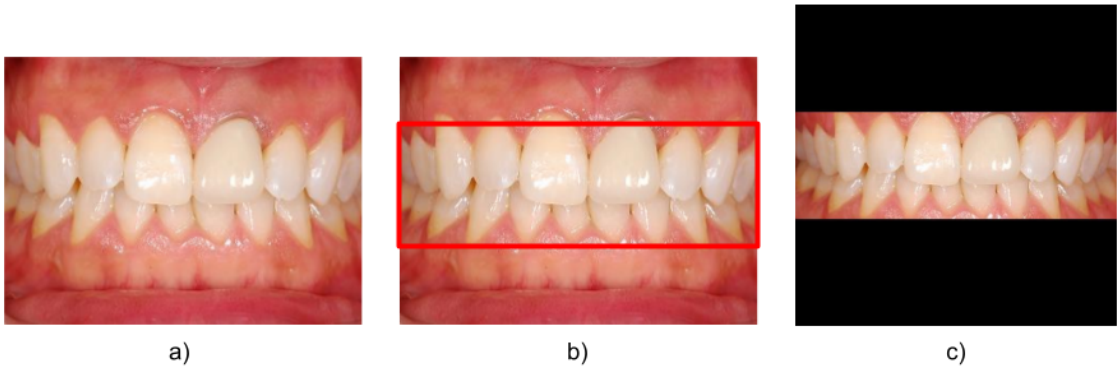

**FIG. 1** Original image from dataset **(a)**, the area we focused on (in the red box) **(b)**, and the cropped version **(c)**.

*1. Segmentation*

The Oral and Dental Spectral Imaging DataBase (ODSI-DB) (Hyttinen et al.) was used for segmentation. The dataset includes 316 images, 215 of which are annotated with manually segmented binary masks. The dataset includes masks with 35 different classes, and images were taken with two different hyperspectral cameras, each with a different number of spectral bands and image dimensions. Of the 215 annotated images, images that didn't contain enamel (a protective coating of the teeth, which was used as an indicator ensuring that teeth were present in the image) were removed, resulting in 158 total images. Additionally, an image with mislabeled annotations was manually removed, resulting in a final total of 157 usable images.

To convert the hyperspectral images in the ODSI-DB dataset into RGB images (in order to emulate a smartphone camera), the reconstruction code and methods provided by Herrera et al. were used. This conversion method follows the work of Magnusson et al. in which the hyperspectral images were converted to the CIE 1931 XYZ color space (Smith and Guild) and subsequently converted to RGB format through mathematical processing.

*2. Classification*

For the classification of calculus, the dataset provided by Park et al. was used to train and test the model. Of the 220 total images, 70% were used for training, 20% for validation, and 10% for testing. Following the methods employed in Park et al.'s study, a bounding box was manually created around the teeth in each image to isolate the area of interest, and everything else appearing in the image was cropped out. This was accomplished using the Roboflow online editing tool ([https://roboflow.com/](https://roboflow.com/)), an open-sourced software used for image labeling and processing. All the images were resized to have dimensions of 640 x 640 while maintaining the aspect ratio of the original image, and the surrounding spaces were padded with black 0s, as shown in Figure 1.

## B. Segmentation Models

A UNet-based model (Ronneberger et al.) was constructed and run on ODSI-DB to segment calculus, but failed to produce any significant results due to the small dataset size and limited processing power available. Previously, Herrera et al. also attempted segmentation of the ODSI-DB dataset, training their model on individual RGB pixels, individual hyperspectral pixels, RGB images, and hyperspectral images. However, the small size of the dataset and the difficulty of segmenting images meant that they were only able to achieve an average accuracy of 52.51% on RGB image classification. Additionally, this accuracy was averaged across all classes, including classes such as attached gingiva and the lip which were predicted with a >90% accuracy but are much easier problems than the segmentation of oral diseases. The low accuracy rate produced by segmentation along with the fact that generating a mobile model would only further decrease the accuracy led us to decide to not pursue segmentation further.

## C. Classification Models and Mobile Application

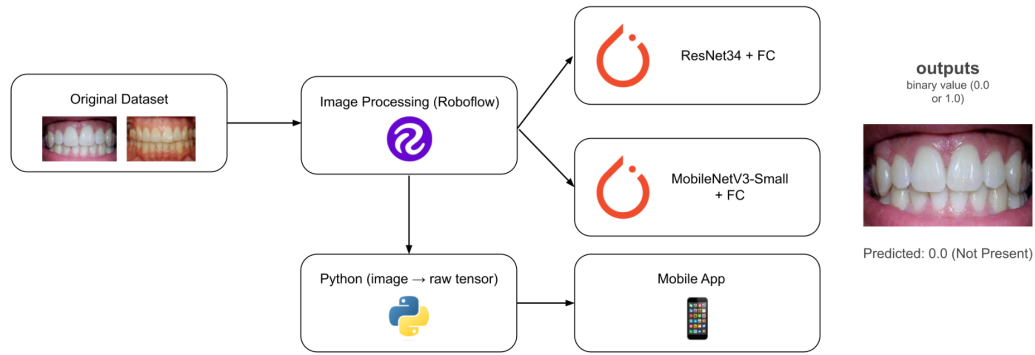

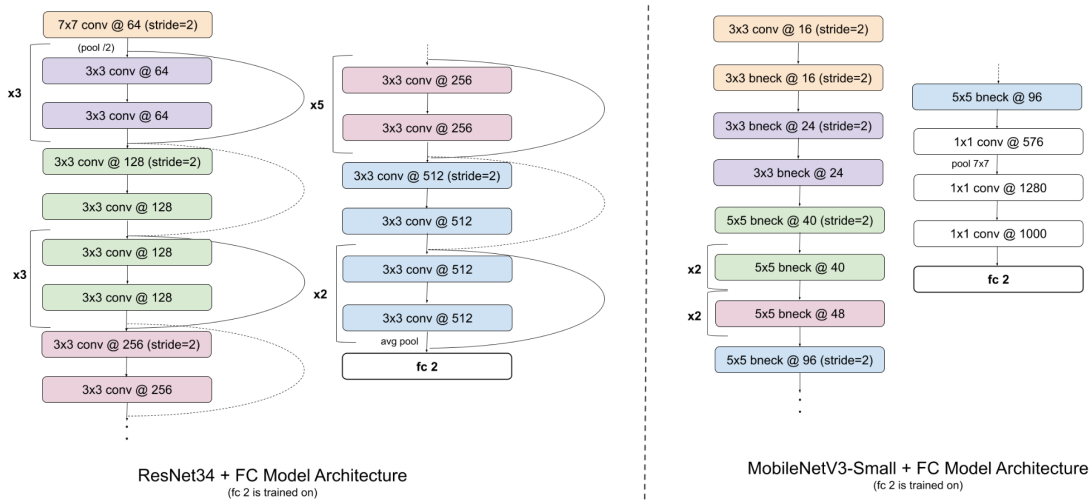

**FIG. 2** Training procedure flowchart and ResNet34 and MobileNet model architectures. Image processing was done manually and then fed into both models, and the preprocessed images were converted to raw tensors before being fed into the app.

A ResNet34 model was constructed using weights pre-trained on the ImageNet database (Deng et al.) and connected to a single Fully Connected layer, changing the model into a binary classification model. Only the last layer was trained to minimize processing time and processing power. The batch size was 128 images for ResNet and the model was trained for 30 epochs. The Adam optimizer (Kingma and Ba) and the Binary Cross-Entropy loss function were used, and the models were trained using PyTorch (Paszke et al.). Other models, such as ResNet152, ResNet18, and ResNet50 were also tested, but ResNet34 produced the highest accuracy and most stability as the validation accuracy would remain constant around a certain value over consecutive epochs.

Similarly, a MobileNetV3-Small model was constructed using weights pre-trained on the ImageNet database (Deng et al.). This model was also connected to an extra Fully Connected layer to do binary classification. However, this model converged after 50 epochs (as shown in Figure 4) with the same loss function and optimizer. The batch size for this model was 32 images.

For each model, hyperparameters, including the batch size, number of epochs, normalization of images, and more were adjusted throughout the tuning process. The Adam optimizer was also compared to the Stochastic Gradient Descent (SGD) optimizer, but the Adam optimizer worked best. The loss function, Binary Cross-Entropy, can be seen in Formula 1. $N$ is the number of images in the dataset, $y_i$ is the $i$-th truth value, and $\hat{y}_i$ is the $i$-th predicted value. In the code, the Cross-Entropy function is used with $N = 2$, resulting in the same function.

$$L = -\frac{1}{N} \sum_{i=1}^{N} [y_i \log(\hat{y}_i) + (1 - y_i) \log(1 - \hat{y}_i)]$$

**FORMULA 1.** The Binary Cross-Entropy Loss formula. Note that log is calculated in base 2 (which is the same as lg).

A simple app was also created to test the model on a mobile device. The mobile application was created with Android Studio using the PyTorch Android Lite package and tested on a Pixel 3a emulated device. The images were preprocessed by Python and loaded into the Android app as raw tensor files due to differences in the way that the images were loaded across both platforms; however, future iterations of the app can remove this step by unifying the image loading process. Screenshots of the mobile app can be seen in Figure 6.

## III. RESULTS

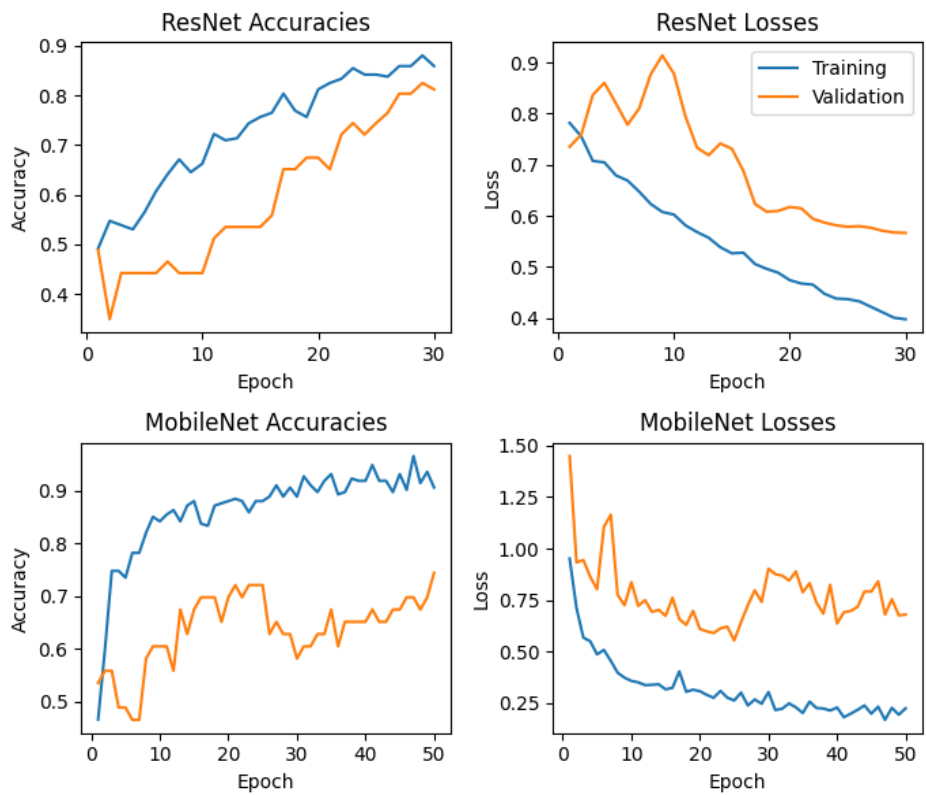

**FIG. 3** ResNet 34 and MobileNetV3-Small accuracies and losses across each epoch. ResNet34 achieved a peak validation accuracy of 82.48% and MobileNet achieved a peak validation accuracy of 74.42%.

Over the course of training the model, the ResNet34 model achieved a maximum accuracy of 88.03% on training data while achieving a maximum accuracy of 82.48% on validation data. MobileNet generated a higher accuracy on training data, with a 96.58% maximum accuracy, but a lower accuracy on validation data, with a 74.42% maximum accuracy. Additionally, the ResNet34 model generated a minimum loss of 0.398 on training data and a minimum loss of 0.567 on validation data. MobileNet received a lower loss on training data than the ResNet34 model, with a loss of 0.169, and a lower loss rate on validation data, with a loss of 0.554.

The ResNet34 model appears to follow a traditional accuracy curve, starting with a lower accuracy which gradually increases across epochs of training. However, the MobileNetV3-Small model's validation accuracy suddenly falls around the 25th epoch (see Figure 3). We hypothesize that this unusual behavior in the MobileNet model's accuracy curve occurred because the model initially trained using one feature in the input images but subsequently found a more suitable feature to learn from, causing the validation accuracy to briefly drop before increasing again towards the end of the graph.

The confusion matrix for both models can also be seen in Figure 4. The MobileNet would more often predict that calculus was present rather than not, but the ResNet produced a more distributed and balanced output.

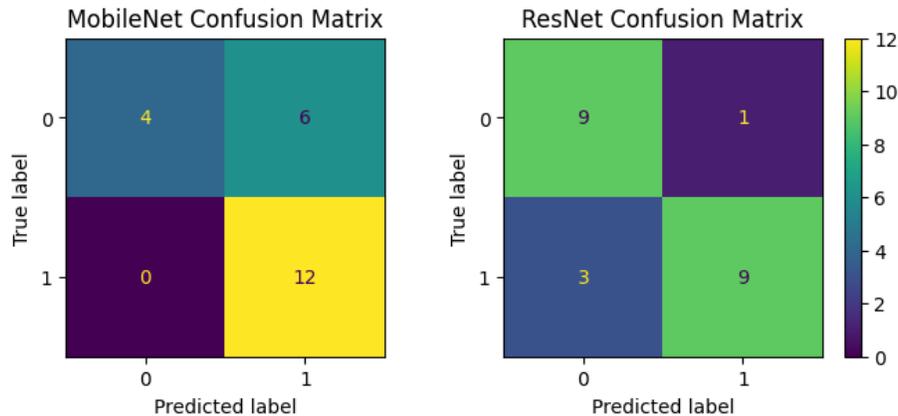

**FIG. 4** Confusion matrices for the ResNet34 and MobileNet models. The ResNet produced a much more balanced distribution and gained a higher accuracy while the MobileNet skewed towards labeling images as containing calculus.

| Names | Accuracy | Recall | F1 score | Precision | MACs |
| --- | --- | --- | --- | --- | --- |
| ResNet34 | 81.82 | 75.00 | 81.82 | 90.00 | 3675117058.0 |
| MobileNetV3-Small | 72.73 | 100.00 | 80.00 | 66.67 | 60127442.0 |

**TABLE 1.** ResNet and MobileNet scores across each category. MobileNet used much fewer operations but ResNet34 achieved a higher accuracy, precision, and f1-score.

The ResNet34 model achieved a higher accuracy, a higher F1 score, and a higher precision rate than the MobileNet model (see Table 1). However, the MobileNet model achieved a higher recall and used fewer multiply-accumulate (MAC) operations. Therefore, the ResNet34 model is more accurate and precise than the MobileNet model but the MobileNet model is more efficient.

| Reference | Name | Accuracy | Image Type | Mobile-based | Architecture |
| --- | --- | --- | --- | --- | --- |
| **Proposed solution** | MobileNetV3-Small | 72.73 | RGB Intraoral | Yes | MobileNet |
| **Proposed solution** | ResNet34 | 81.82 | RGB Intraoral | Yes | ResNet |
| Park et al. | Parallel 1D Conv + Shortcut | 74.54 | RGB Intraoral | No | Other/Custom |
| Mao et al. | GoogLeNet | 94.97 | Periapical Radiograph | No | GoogLeNet |
| Liu et al. | Mask R-CNN | 95.10 | RGB Intraoral | Yes | Mask R-CNN |
| Prajapati et al. | VGG16 | 88.46 | RadioVisiography (RVG) X-ray images | No | VGG16 |

**TABLE 2.** Comparison of previous and proposed solutions. In contrast to previous works, the proposed solutions utilize both RGB images and lightweight machine learning architectures capable of running on a smartphone, increasing accessibility while maintaining comparable accuracy.

Our model achieved a comparable accuracy compared to other state-of-the-art models with higher processing power and larger datasets. Other than Liu et al., our proposed models are the only models to use RGB intraoral images while being mobile based. However, unlike Liu et al., our mobile app does not require an internet connection, making it more accessible to patients in countries with trouble accessing the internet - and in the future, we hope to increase the accuracy to match theirs with a larger dataset and more processing power.

### A. Model Comparison on Mobile App

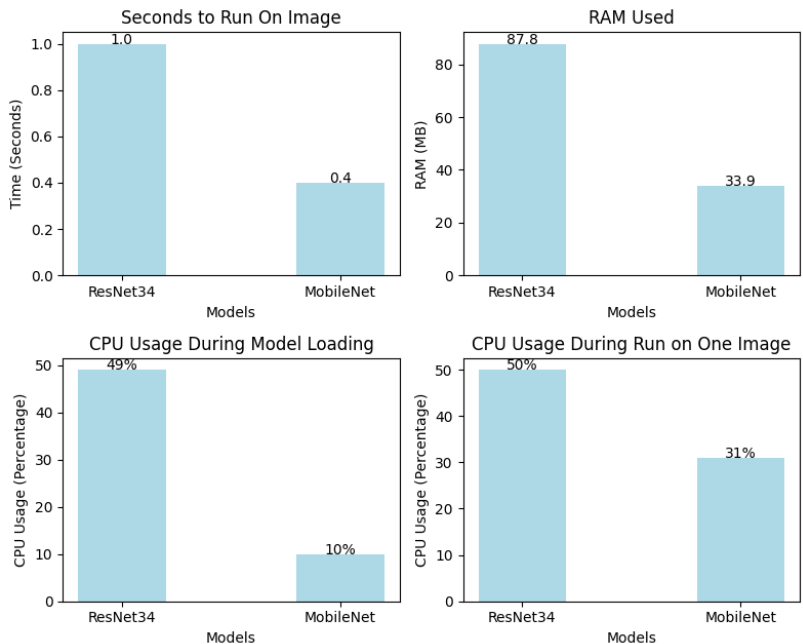

**FIG. 5** ResNet and MobileNet efficiency on a mobile app. MobileNet outperformed ResNet in all four graphical comparisons, showing its efficiency compared to the ResNet model.

While running both models on a Google Pixel 3a emulator, the MobileNet model was much faster than the ResNet34 model (see Figure 5). In addition, the MobileNet model used less RAM to run and used less CPU during model loading and classifying an image (see Figure 5). Overall, the MobileNet model outperformed the ResNet34 model in terms of speed and space efficiency.

### B. Mobile App

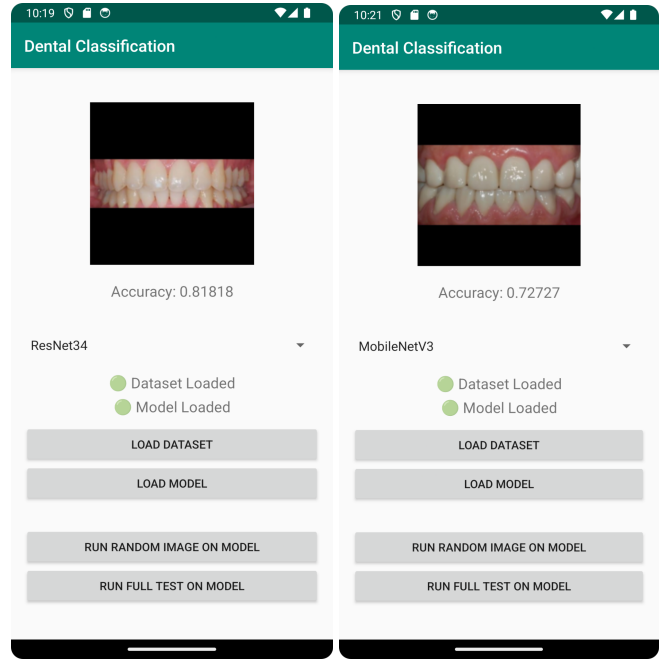

**FIG. 6** Using the mobile app, a random image on ResNet **(a)** and a random image on MobileNet **(b)**. The app produced the same accuracy using both MobileNet and ResNet as while running the original models.

Using the mobile app, we were able to achieve the same accuracy as by running our original models, thus proving the feasibility of incorporating models like ours into apps on mobile devices.

## IV. CONCLUSION

Due to the global prevalence of oral diseases, the need for fast, accessible oral diagnostic care is necessary. Most state-of-the-art research using neural networks to classify oral disease does not consider accessibility to patients as a factor, as most models use X-rays, not RGB images, and are incapable of being run without an internet connection or on low-powered devices such as smartphones. To rectify this, we were able to construct a ResNet model and a MobileNet model which were able to produce accuracies of 81.82% and 72.73%, respectively, as well as a mobile app capable of running both of these models on low-end devices.

However, our research had some limitations, which can be rectified in future work. Model training needs to be done with a larger dataset, which would allow our model to generalize better over a larger variety of images. Longer training times and increased processing power can also be used to achieve a higher detection accuracy, especially by training on all of the layers in the model instead of transfer learning. Our pipeline also doesn't contain a method for automating the image processing step, so a bounding box detection algorithm can be used to automatically crop out teeth from images, allowing for a completely standalone app in which all the image processing occurs right before the model is run.

In the future, our models and app can be used in regions with low internet infrastructure to provide residents with low-cost dental diagnoses. Furthermore, the success of the mobile app shows how it can be expanded in the future to allow for the classification of many dental diseases by running directly from pictures taken by the user, thereby increasing the scope of oral diagnostic care to those who do not have access to specialized equipment such as X-ray machines. The app can integrate with other metrics to allow for better detection accuracy (such as patients' health history) and can allow the user to send information to the doctor if a disease is detected for further evaluation.

We made dental diagnosis more accessible by building a standalone smartphone app that does not require an internet connection to run. The app is able to run our MobileNet and ResNet models that classified whether the user-provided RGB intraoral images contained calculus or not. Our accuracies were comparable to other state-of-the-art models while also being lightweight and usable in a mobile application capable of running on low-end devices.

## ACKNOWLEDGEMENTS

We would like to thank our track instructor, Shailja, as well as our teaching assistants, Satish Kumar and Arthur Caetano, for all their help, guidance, and support with our research and this paper. We also thank Dr. Lina Kim and the Summer Research Academies program for providing the opportunity and facilities to conduct research at the University of California, Santa Barbara campus.

## AUTHOR CONTRIBUTION STATEMENT

A.G., J.L., and A.M. conceived the research topic. A.G. built and trained the models, and constructed the mobile app. J.L. constructed the graphs. A.M. created the figures. All authors contributed to the manuscript.